\pdfoutput=1

\documentclass[11pt]{article}

\usepackage{EMNLP2022} %

\usepackage{times}
\usepackage{latexsym}
\usepackage{amsmath}
\usepackage{amssymb}
\usepackage{array}
\usepackage{graphicx}
\usepackage{subfigure}
\usepackage{multirow}
\usepackage[encapsulated]{CJK}
\usepackage{float}
\usepackage{mathrsfs}
\usepackage{mathtools}
\usepackage{amsopn}
\usepackage{url}
\usepackage{soul}
\usepackage{longtable}
\usepackage{arydshln}
\usepackage{pgfplots}
\pgfplotsset{compat=1.15}

\usepackage{enumerate}

\usepackage{amssymb}
\usepackage{lineno}   
\usepackage{subfigure} 
\usepackage{multirow}
\usepackage{arydshln}
\usepackage{enumitem}
\usepackage{amsmath}
\usepackage{footnote}
\usepackage{bm}
\usepackage{graphicx}
\usepackage{color}
\usepackage{stfloats}
\makesavenoteenv{table}
\usepackage{tablefootnote}
\usepackage[T1]{fontenc}

\usepackage[utf8]{inputenc}

\usepackage{microtype}

\usepackage{inconsolata}

\title{Distill the Image to Nowhere: Inversion Knowledge Distillation for Multimodal Machine Translation}
\author{Ru Peng\textsuperscript{1$\ast$}, Yawen Zeng\textsuperscript{2}\thanks{Both authors contributed equally to this research.}, Junbo Zhao\textsuperscript{1}\thanks{Corresponding author.} \\
  \textsuperscript{1}Zhejiang University, Zhejiang, China \\
  \textsuperscript{2}Tencent WeChat, Shenzhen, China \\
  \texttt{pengru709909347@gmail.com, yawenzeng11@gmail.com, j.zhao@zju.edu.cn}
}

\begin{document}
\maketitle
\begin{abstract}
Past works on multimodal machine translation (MMT) elevate bilingual setup by incorporating additional aligned vision information.
However, an \emph{image-must} requirement of the multimodal dataset largely hinders MMT's development --- namely that it demands an aligned form of [image, source text, target text].
This limitation is generally troublesome during the inference phase especially when the aligned image is not provided as in the normal NMT setup.
Thus, in this work, we introduce IKD-MMT, a novel MMT framework to support the \emph{image-free} inference phase via an \emph{inversion knowledge distillation} scheme.
In particular, a multimodal feature generator is executed with a knowledge distillation module, which directly generates the multimodal feature from (only) source texts as the input.
While there have been a few prior works entertaining the possibility to support image-free inference for machine translation, their performances have yet to rival the image-must translation.
In our experiments, we identify our method as the first image-free approach to comprehensively rival or even surpass (almost) \emph{all} image-must frameworks, and achieved the state-of-the-art result on the often-used Multi30k benchmark\footnote{Our code and data are available
at: \url{https://github.com/pengr/IKD-mmt/tree/master.}}.
\end{abstract}

\begin{figure}[!t]
\centering
{\includegraphics[width=1.\linewidth]{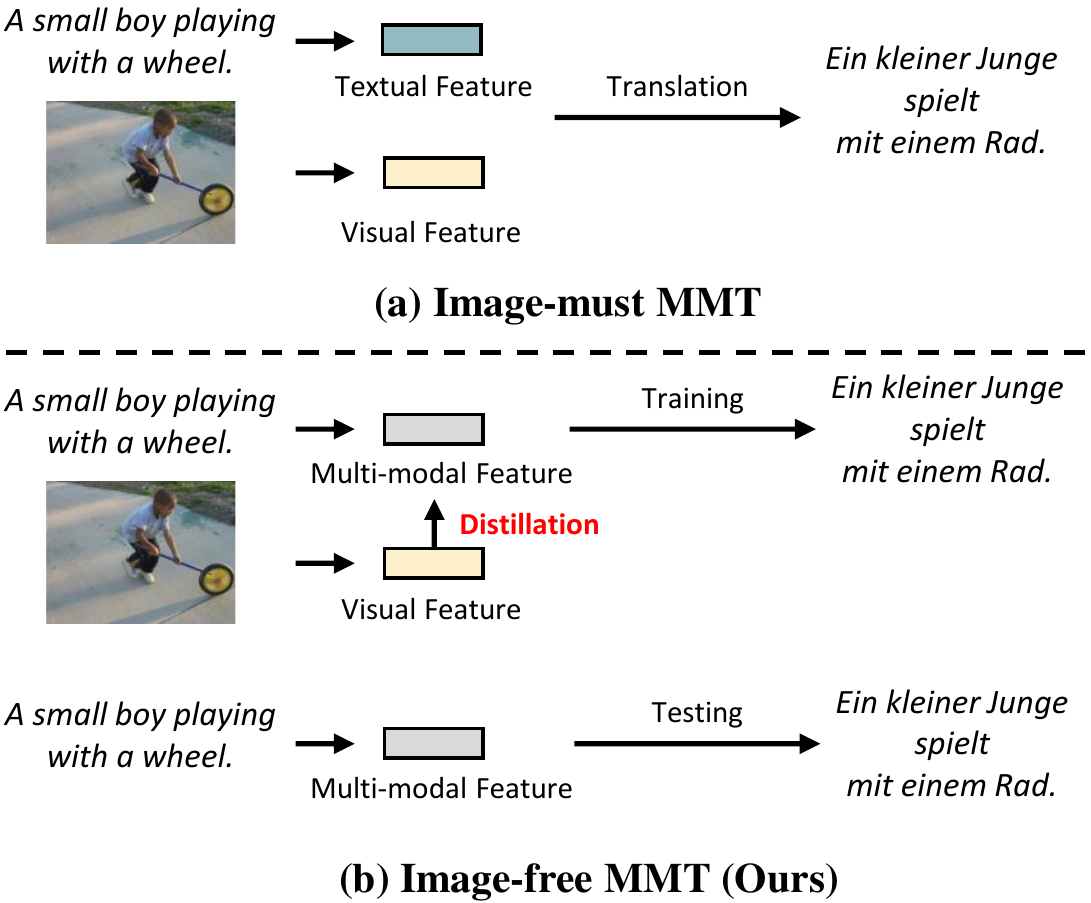}}
\centering
\caption{Examples of Image-must MMT (a), and our Image-free MMT (b). During testing, our IKD-MMT does not require the image as input.}
\label{fig:Fig1}
\end{figure}

\section{Introduction}
\label{sec:intro}
Multimodal machine translation (MMT) is an worthy task of elevating text-only translation by introducing additional image modality \cite{specia-etal-2016-shared,elliott-etal-2017-findings,barrault-etal-2018-findings}.
Existing works mostly focus on the fusion and alignment of images and texts to improve MMT \cite{calixto-etal-2017-doubly,ive-etal-2019-distilling,yin-etal-2020-novel}, that they have managed to concept-prove the effectiveness of the aligned visual information.
Nevertheless, the strict triplet data form of the dataset, in both the training and inference phases, has disabled the MMT model to generalize further.
In particular, if we consider using an MMT model to conduct translation for the normal bilingual text translation as in the NMT setup, one must provide the aligned images during inference.
And unfortunately, this is not often feasible.
This general comparison between image-free and image-must schemes is visually illustrated in Figure \ref{fig:Fig1}(a).
In hindsight, the quantity and quality of attached images become a bottleneck towards the development of MMT, as acquiring such resources can be scarce and expensive (e.g. Multi30K \cite{elliott-etal-2016-multi30k}).

Indeed, there have been a few attempts to resolve the image-must limitation.
For instance, \citet{elliott-kadar-2017-imagination} present a multi-task learning model for MMT where they rely on an auxiliary visual grounding task to obtain the visual feature. \citet{Zhang2020Neural} introduce an image retrieval paradigm to find topic-related images from a small-scale dataset. Further, \citet{long-etal-2021-generative} attempts to utilize a set of generative adversarial networks to obtain an imaginary vision feature.
We may posit that a (nearly) common ground for such image-free frameworks is to learn and further obtain a generated visual feature representation without the actual image data provided during inference.
However, none of the aforementioned works has managed to consistently reach the performance of the image-must counterpart.
In this work, we hypothesise that this can be caused by the inferior representation learned, insufficient visual distribution coverage, improper multimodal fusion stage~\cite{caglayan-etal-2017-lium,arslan2018doubly,helcl-etal-2018-cuni,calixto-liu-2017-incorporating}, and/or lacked training stability, etc.

In this work, we intend to take a thorough exploration towards this line.
As Shown in Figure~\ref{fig:Fig1}(b), unlike prior works solely targeting visual feature generation and/or relying on later stages of fusion, our approach directly generates a \textbf{multimodal feature} using only the source text input.
We enable this by proposing an inverse knowledge distillation mechanism employing pre-trained convolutional neural networks (CNN).
From our experiments, we find that this architectural choice has notably enhanced the training stability as well as the final representation quality.
To this end, we introduce the IKD-MMT framework, an image-free framework that systematically rivals or outperforms the image-must frameworks.
To set up the inverse knowledge distillation flow, we incorporate dual CNNs with inverted data feeding flow.
Of the two, the teacher network receives the pre-trained weights while the student CNN is trained from scratch aiming to provide a high-quality multimodal feature space by incorporating both inter-modal and intra-modal distillations.

Our contributions are summarized as follows:

\textbf{i.} IKD-MMT framework is the first method that systematically rivals or even outperforms the existing image-must frameworks, which fully demonstrates the feasibility of the image-free concept;

\textbf{ii.} We pioneer the exploration of knowledge-distillation combined with the pre-trained models in the regime of MMT, as well as the multimodal feature generation. We posit that these techniques have shed some light on the representation learning and training stability of MMT.


\begin{figure*}[!t]
\centering
\setlength{\belowcaptionskip}{-12pt}
{\includegraphics[width=0.9\linewidth]{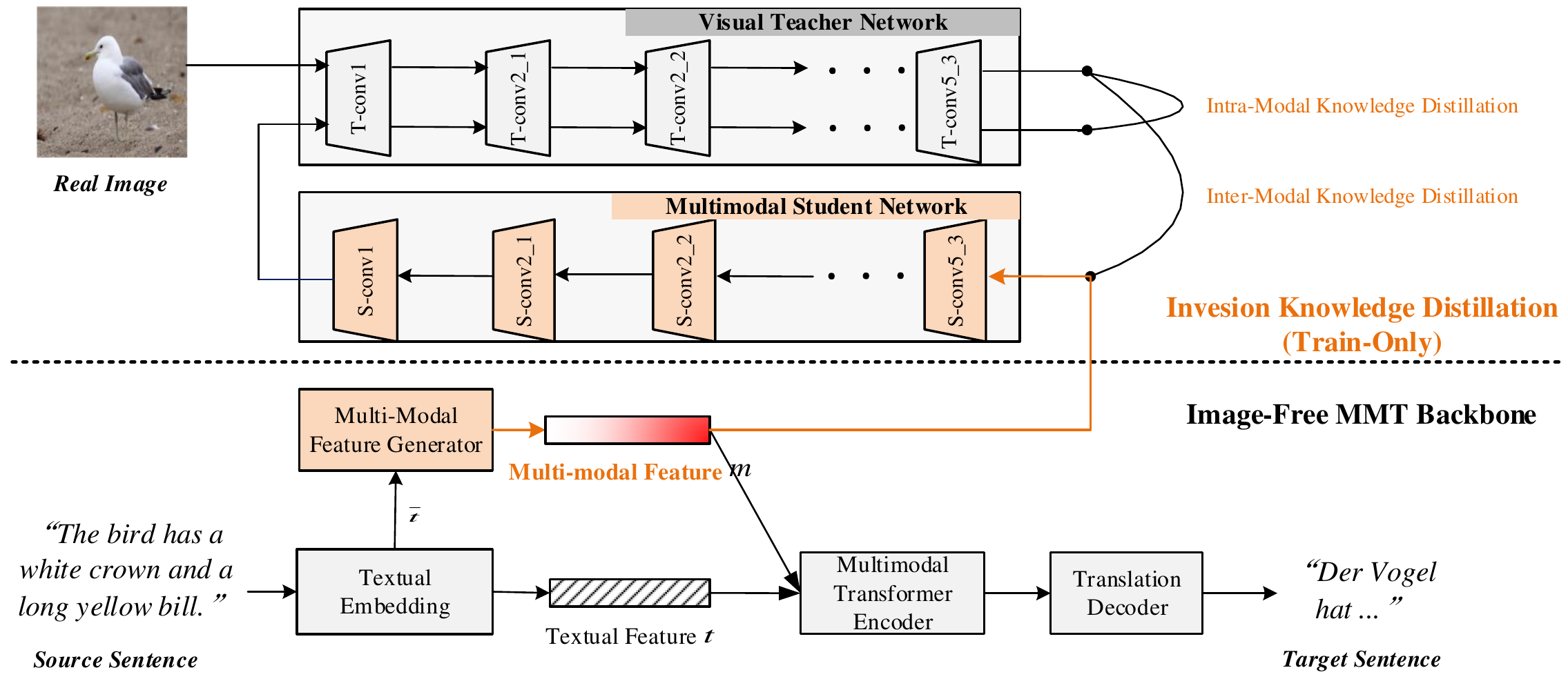}\label{fig2}}
\centering
\caption{The framework of our IKD-MMT model.
The multimodal feature generator, multimodal student network and visual teacher network are the most critical modules, which help break the dataset constraints of image-must.
}
\label{fig:Fig2}
\end{figure*}

\section{Related Work}
\subsection{Multi-modal Machine Translation}
As an intersection of multimedia and neural machine translation (NMT), MMT has drawn great attention in the research community.
Technically, existing methods mainly focus on how to better integrate visual information into the framework of NMT.
1) \citet{calixto-etal-2017-doubly} propose a doubly-attentive decoder to incorporate two separate attention over the source words and visual features.
2) \citet{ive-etal-2019-distilling} propose a translate-and-refine approach to refine draft translations by visual features.
3) \citet{yao-wan-2020-multimodal} propose the multimodal Transformer to induce the image representations from the text under the guide of image-aware attention.
4) \citet{yin-etal-2020-novel} employs a unified multimodal graph to capture various semantic interactions between multimodal semantic units.

However, the quantity and quality of the annotated images limit the development of this task, which is scarce and expensive.
In this work, we aim to perform the MMT in an image-free manner, which has the ability to break data constraints.

\subsection{Knowledge Distillation}
Knowledge distillation (KD) \cite{BucilucoModel2006,hinton2015distilling} aims to use a knowledge-rich teacher network to guide the parameter learning of the student network.
In fact, KD has been investigated in a wide range of fields.
\citet{romero2014fitnets} transfer knowledge through an intermediate hidden layer to extend the KD.
\citet{yim2017gift} define the distilled knowledge to be transferred in terms of flow between layers, which is calculated by the inner product between features from two layers.
In the multimedia field, \citet{gupta2016cross} first introduce the technique that transfers supervision between images from different modalities.
\citet{yuan2018text} propose the symmetric distillation networks for the text-to-image synthesis task.

Inspired by these pioneering efforts, our IKD-MMT framework is intents to take full advantage of KD to generate a multimodal feature to overcome triplet data constraints.

\section{IKD-MMT Model}
As illustrated in Figure \ref{fig:Fig2}, the proposed framework consists of two components:
an image-free MMT backbone and a multimodal feature generator.

\subsection{Image-Free MMT Backbone}
Given a source sentence $X\text{=}\left( {{x}_{1}},\ldots ,{{x}_{I}} \right)$, each token ${x}_{i}$ is mapped into a word embedding vector ${{E}_{{{x}_{i}}}}\in {{\mathbb{R}}^{{{d}_{w}}}}$ through the textual embedding with position encoding \cite{ gehring-2017-convolutional}.
${d_{w}}$ and $t=\left( {{E}_{{{x}_{1}}}},\ldots ,{{E}_{{{x}_{I}}}} \right)$ are the word embedding dimension and the textual feature, respectively.

Then, we feed the text feature $t$ together with the multimodal feature $m$ (detail in Section 3.2.1) into the multimodal transformer encoder \cite{yao-wan-2020-multimodal}.
In the multimodal encoder layer, we cascade the multimodal feature $m$ and the text feature $t$ to reorganize a new multimodal feature $\widetilde{x}$ as the query vector:
\begin{gather}
\widetilde{x}\text{=}\left[ t;m{{W}^{m}} \right]\in {{\mathbb{R}}^{(I+P)*d}}\label{eq:1},
\end{gather}

where $I$ is the length of source sentence, and $P$ is the size of multimodal feature. Here, we can understand this modal fusion from the perspective of nodes and graphs.
If we treat each source token as a node, each region of the multimodal feature can also be regarded as a pseudo-token and added to the source token graph for modal fusion.
The key and value vectors are preserved as the text feature $t$, and the multimodal encoder layer is calculated as follows:
\begin{gather}
{{c}_{k}}=\underset{i=1}{\overset{I}{\mathop{\sum }}}\,{{\tilde{\alpha }}_{ki}}\left( {{t}_{i}}{{W}^{V}} \right),\label{eq:2}\\
{{\tilde{\alpha }}_{ki}}=softmax\left( \frac{\left( {{{\tilde{x}}}_{k}}{{W}^{Q}} \right){{\left( {{t}_{i}}{{W}^{K}} \right)}^{\top }}}{\sqrt{d}} \right).\label{eq:3}
\end{gather}

In this paper, we directly adopt the Transformer decoder\footnote{For details, please refer to the original paper.} \cite{vaswani2017attention} for translation.
Given a target sentence $Y\text{=}\left( {{y}_{1}},\ldots ,{{y}_{J}} \right)$, our framework outputs the predicted probability of the target word ${{y}_{j}}$ as follow:
\begin{gather}
p\left( {{y}_{j}}| {{y}_{<j}},\text{X, }m \right)\propto \exp \left( {{W}^{h}}H_{j}^{L}+{{b}^{h}} \right),\label{eq:4}
\end{gather}
where $H_{j}^{L}$ represents the top output of the decoder at $j$-th decoding time step, ${{W}^{h}}$ and ${{b}^{h}}$ are learnable multi-layer perceptrons, and $\exp () $ is a Softmax layer.

\subsection{Multimodal Feature Generation}
\subsubsection{Preliminaries}
In this part, we introduce the frame, symbol definitions and task goal of multimodal feature generation in advance.

The frame is composed of a multimodal feature generator $F$, a visual teacher model $T$ and a multimodal student model $S$.
The detailed architecture of each module is shown in Table \ref{table_VII} of the appendix.
The model parameters of $S$ are denoted as ${{\theta }^{s}}$.
When the global text feature $\overline{t}$ is fed into $S$, the hidden representation produced by the $l$-th layer is denoted as $\varphi _{l}^{S}\left( \overline{t},\theta _{l}^{s} \right)$.
The $F$ outputs a multimodal feature $m$, and the $S$ produces an inverse feature ${{I}_{s}}$ after the \textit{S-conv1} layer.
The real image and the inverse feature are $\left\{ {{I}_{s}},{{I}_{r}} \right\}\in {{\mathbb{R}}^{m*n*3}}$.
Given a feature $I$ as input, the hidden representation produced by the $l$-th layer of $T$ is denoted as $\varphi _{l}^{T}(I)$.

Our goal is to generate multimodal features from the source text to break the image-must restriction in testing.
The visual perception of this multimodal feature is extracted from the visual distillation of the teacher-student model, while the textual semantic of that is derived from the text translation of the input text.

\subsubsection{Multimodal Feature Generator}
First, we simply adopt an average pooling to transform all word embedding vectors into global textual features, which are proven to carry the overall word senses in \cite{zhang2010understanding}.
\begin{gather}
\overline{t}\text{ = }{}^{1}/{}_{I}\sum\limits_{i=1}^{I}{{{E}_{{{x}_{i}}}}}.\label{eq:5}
\end{gather}

Then, the global text feature $\overline{t}$ is serially transported into the multimodal feature generator to compute a multimodal feature $m$:
\begin{gather}
m=\text{unpool(}{{W}^{t}}\overline{t}\text{)}.\label{eq:6}
\end{gather}

Among them, the FC layer ${W}^{t}$ projects the global text feature $\overline{t}$ into the image space.
The following average unpooling computes a high-dimensional multimodal feature map from the low-dimensional latent vector.
The dimension of $m\in{{\mathbb{R}}^{P*2048}}$ is the same as that of the last convolutional activation of the teacher model.
Notably, the textual semantics of multimodal features are modelled from the global textual context supervised by the text translation.

\subsubsection{Inversion Knowledge Distillation}
The inversion knowledge distillation transfers the visual perception from the teacher model $T$ to the student model $S$, and in-depth interacts with textual semantics in the multimodal feature generator.
To synthesize an information-rich multimodal feature, we formulate a novel dual distillation paradigm consisting of inter-modal (IrM-KD) and intra-modal (IaM-KD) knowledge distillations.

\textbf{IrM-KD}:
The IrM-KD direct the student model $S$ to extract the vital visual information from the source text, thereby bridging the inter-modal semantics of the text and the real image.
Specifically, given the real image ${I}_{r}$, the teacher model $T$ generates a visual representation $\varphi _{l}^{T}({{I}_{r}})$ in each layer $l$.
Meanwhile, the $S$ produces a inverse hidden representation $\varphi _{l+1}^{S}\left( \overline{t},\theta _{l}^{s} \right)$ in next layer $l+1$.
The paired representations $\varphi _{l}^{S}\left( \overline{t},\theta _{l}^{s} \right)$ and $\varphi _{l}^{T}({{I}_{r}})$ with identical dimension entail the same-level latent concepts.
We present the IrM-KD loss by the discrepancy among these two representations and an auxiliary regularization term:
\begin{eqnarray}
Los{{s}_\text{IrM}}\text{=}\sum\limits_{l}{{{\left\| \varphi _{l}^{T}\text( {{I}_{r}} \text)\text{-}\varphi _{l+1}^{S}\text( \overline{t};\theta _{l}^{s} \text) \right\|}_{2}}}\text{+}{{\left\| {{I}_{r}}\text{-}{{I}_{s}} \right\|}_{2}},\label{eq:7}
\end{eqnarray}
where the $L_{2}$ norm ${{\left\| \right\|}_{2}}$ is used to measure the similarity of two vectors. The regularization term ${{\left\| {{I}_{r}}\text{-}{{I}_{s}} \right\|}_{2}}$ indicates the image space loss, which is the fundamental constraint for the $S$ to learn the distribution of the real image.

\textbf{IaM-KD}:
The IaM-KD constrains the student model $S$ to learn the visual perception of images via the inverse feature, thus relieving the intra-modal gap between the inverse feature with the real image.
Specifically, we fed the inverse feature ${I}_{s}$ into the teacher model $T$ to gain the teacher’s cognition for it --- a pseudo visual representation $\varphi _{l}^{D}\left( {{I}_{s}} \right)$.
Then, to encourage the student model profoundly learn the distribution of images, we narrow the divergence between the $\varphi _{l}^{D}\left( {{I}_{s}} \right)$ and its coupled visual representation $\varphi _{l}^{T}({{I}_{r}})$.
So that, the IaM-KD loss is defined as the combination of the above divergence and the image space loss:
\begin{gather}
Los{{s}_{\text{IaM}}}\text{=}\sum\limits_{l}{{{\left\| \varphi _{l}^{T}\text( {{I}_{r}} \text)\text{-}\varphi _{l}^{D}\text( {{I}_{s}} \text) \right\|}_{2}}}\text{+}{{\left\| {{I}_{r}}\text{-}{{I}_{s}} \right\|}_{2}}.\label{eq:8}
\end{gather}

Compare with T2I synthesis works \cite{reed2016generative,zhang2017stackgan,xu2018attngan}, we are dedicated to aidding text translation through inter-modal and intra-modal bi-visual distillation.
By doing so, our generated multimodal feature focuses more on the text-image alignment and fusion, but not only the authenticity of image.

\subsection{Objective function}
During the training phase, we optimize the proposed IKD-MMT model end-to-end by the text translation loss and the inversion distillation loss:
\begin{gather}
J\text( \theta ,{{\theta }_{s}} \text)\text{=}{{J}_{\text{trans}}}\text( \theta ,{{\theta }_{s}} \text)\text{+}Los{{s}_{\text{IrM}}}\text{+}Los{{s}_{\text{IaM}}}.\label{eq:9}
\end{gather}

Wherein, the translation loss over the training dataset $\mathcal{D}$, not only bridges the relevance of the source and target texts, but also models the text semantics of multimodal features:
\begin{gather}
{{J}_{\text{trans}}}\text( \theta ,{{\theta }_{s}} \text)\text{=-}\underset{D}{\mathop{\sum }}\,\underset{J}{\mathop{\sum }}\,\log p\text( {{y}_{j}}|{{{y}_{<j}}},X,m \text).\label{eq:10}
\end{gather}

In the testing phase, the trained multimodal feature generator is capable to generate rich features to embed into the MMT backbone, thus getting rid of the image-must constraints.

\begin{table*}[!t]
\centering
\renewcommand{\arraystretch}{0.85}
\setlength{\tabcolsep}{1pt}
\caption{BLEU (``B'') and METEOR (``M'') scores of EN-DE and EN-FR tasks.
Encouragingly, our IKD-MMT as an image-free MMT model outperforms almost all MMT systems, and even rivals the SOTA image-must systems.
$\ddagger$/$\dagger$ mark statistically significant variations for BLEU ($p$-value $<$ 0.01/0.05) as compared to the Transformer.}
\label{table_I}
\centering
{\begin{tabular}{l|cccccc|cccc}
\hline
\hline
\multicolumn{1}{c|}{\multirow{3}*{\textbf{Systems}}} & \multicolumn{6}{c}{\textbf{EN-DE}} & \multicolumn{4}{c}{\textbf{EN-FR}} \\
\cline{2-11}
& \multicolumn{2}{c}{\textbf{Test2016}} & \multicolumn{2}{c}{\textbf{Test2017}} & \multicolumn{2}{c}{\textbf{MSCOCO}} & \multicolumn{2}{c}{\textbf{Test2016}} & \multicolumn{2}{c}{\textbf{Test2017}} \\
\cline{2-11}
& B & M & B & M & B & M & B & M & B & M \\
\hline
\multicolumn{11}{c}{\textit{Image-must MMT Systems}}\\
\hline
NMT$_{\text{SRC+IMG}}$\cite{calixto-etal-2017-doubly} & 36.5 & 55.0 & - & - & - & - & - & - & - & - \\
IMG$_D$\cite{calixto-liu-2017-incorporating} & 37.3 & 55.1 & - & - & - & - & - & - & - & - \\
Fusion-conv\cite{caglayan-etal-2017-lium} & 37.0 & 57.0 & 29.8 & 51.2 & 25.1 & 46.0 & 53.5 & 70.4 & 51.6 & 68.6 \\
Trg-mul\cite{caglayan-etal-2017-lium} & 37.8 & 57.7 & 30.7 & 52.2 & 26.4 & 47.4 & 54.7 & 71.3 & 52.7 & 69.5 \\
VAG-NMT\cite{zhou-etal-2018-visual} & - & - & 31.6 & 52.2 & 28.3 & 48.0 & - & - & 53.8 & 70.3 \\
DS-SUM-L2\cite{caglayan2019multimodal} & 39.4 & 58.7 & 32.6 & 52.9 & - & - & 60.7 & 76.0 & 54.2 & 71.0 \\
Del+obj\cite{ive-etal-2019-distilling} & 38.0 & 55.6 & - & - & - & - & 59.8  & 74.4 & - & - \\
Multimodal\cite{yao-wan-2020-multimodal} & 38.7 & 55.7 & - & - & - & - & - & - & - & - \\
GMNMT\cite{yin-etal-2020-novel} & 39.8 & 57.6 & 32.2 & 51.9 & 28.7 & 47.6 & 60.9 & 74.9 & 53.9 & 69.3 \\
DCCN\cite{lin2020dynamic} & 39.7 & 56.8 & 31.0 & 49.9 & 26.7 & 45.7 & 61.2 & 76.4 & 54.3 & 70.3 \\
Gumbel-att\cite{liu2021gumbel} & 39.2 & 57.8 & 31.4 & 51.2 & 26.9 & 46.0 & - & - & - & - \\
OVC+${L}_{m}$\cite{wang2021efficient} & - & - & 32.3 & 52.4 & 28.9 & 48.1 & - & - & 54.1 & 70.5\\
Gated Fusion\cite{wu2021good} & 41.96 & - & 33.59 & - & 29.04 & - & 61.69 & - & 54.85 & - \\
RMMT\cite{wu2021good} & 41.45 & - & 32.94 & - & 30.01 & - & 62.1 & - & 54.39 & - \\
\hline
\multicolumn{11}{c}{\textit{Image-free MMT Systems}}\\
\hline
Transformer\cite{vaswani2017attention} & 37.6 & 55.3 & 31.7 & 52.1 & 27.9 & 47.8 & 59.0 & 73.6 & 51.9 & 68.3 \\
Multitask\cite{elliott-kadar-2017-imagination} & 36.8 & 55.8 & - & - & - & - & - & - & - & - \\
VMMT$_\text{F}$\cite{calixto-etal-2019-latent} & 37.7 & 56.0 & 30.1 & 49.9 & 25.5 & 44.8 & - & - & - & - \\
UVR-NMT\cite{Zhang2020Neural} & 36.94 & - & 28.63 & - & - & - & 57.53 & - & 48.46 & - \\
ImagiT \cite{long-etal-2021-generative} & 38.5 & 55.7 & 32.1 & 52.4 & 28.7 & 48.8 & 59.7 & 74.0 & 52.4 & 68.3 \\
\hdashline
\multirow{2}*{\textbf{IKD-MMT (Ours)}} & \textbf{41.28}$\ddagger$ & \textbf{58.93} & \textbf{33.83}$\dagger$ & \textbf{53.21} & \textbf{30.17} & \textbf{48.93} & \textbf{62.53}$\dagger$ & \textbf{77.20} & \textbf{54.84}$\dagger$ & \textbf{71.87} \\
& ±0.3 & ±0.20 & ±0.10 & ±0.26 & ±0.14 & ±0.08 & ±0.25 & ±0.18 & ±0.50 & ±0.34 \\
\hline
\hline
\end{tabular}}
\end{table*}

\section{Experiment}
\subsection{Setup}
\quad \textbf{Datasets}
We conduct experiments on the Multi30K benchmark \cite{elliott-etal-2016-multi30k}.
The training and validation sets contain 29,000 and 1,014, respectively.
We report the results of the Test2016, Test2017 and ambiguous MSCOCO test sets.
We directly use the preprocessed sentences\footnote{https://github.com/multi30k/dataset} and apply the BPE \cite{sennrich2016neural} with 10K merge operations to segment words into sub-words, which build a shared vocabulary of 9,712 and 9,544 tokens for EN-DE and EN-FR translation tasks.

\textbf{Settings}
We follow all model settings of \cite{wu2021good}, such as the Transformer-Tiny configuration for anti-overfitting in small datasets.
4-gram case-insensitive BLEU \cite{papineni2002bleu} and METEOR \cite{denkowski-lavie-2014-meteor} are used as evaluation metrics.
All models are run three times and report the average results.

\subsection{Main Results}
\quad \textbf{EN-DE Translation Task}
Table \ref{table_I} reports the performance of all MMT baselines on the En-DE task. Comparing all systems, we draw the following interesting conclusions:

First, the IKD-MMT significantly surpasses all image-free MMT systems on five test sets.
These improvements demonstrate that
a) our model can effectively embed multimodal semantics during the training and guide the translation via multimodal features among the image-free testing phase,
b) benefiting from the informative richness and stable generation of multimodal features, our method is a more robust way to break data constraints.

Second, the image-must MMT systems generally exceed their image-free counterparts, showing the efficacy of additional images for translation.

Finally, encouragingly, our image-free MMT model not only overbeats almost all image-must MMT systems, but even rivals the SOTA image-must MMT.
We speculate that these noticeable gains stem from the IKD-MMT's strong ability to fuse the text semantics and visual perception, and generate text-related visual representation, under the dual supervision of text translation and visual distillation.

\textbf{EN-FR Translation Task}
We also conduct experiments on the EN-FR task.
Our IKD-MMT still outperforms the compared baselines in Table \ref{table_I}.
This verifies the robustness and generality of our model in various language scenarios.
\begin{table*}[!t]
\centering
\caption{Ablation results for diverse distillation variants on the EN-DE task.
The \textit{base} row denotes the IKD-MMT in Table \ref{table_I}, and ``-'' means to retain the setting of the \textit{base} row.
Avg.B and Avg.M indicate the BLEU and METEOR scores of the three test sets
}
\label{table_II}
\centering
\renewcommand{\arraystretch}{0.9} 
\setlength{\tabcolsep}{2.5pt}
{\begin{tabular}{c|cccc|cc}
\hline
\hline
& \textbf{Sim. Func.} & \textbf{Dist. Gran.} & \textbf{CNN Back.} & \textbf{Dist. Loss} & Avg.B & Avg.M \\
\hline
\textit{base} & $L_2$ & Model & ResNet50 & (IrM-KD+IaM-KD) loss & \textbf{35.09} & \textbf{53.69} \\
\hline
\multirow{4}*{(A)} & $L_1$ & - & - & - & 34.60 (-0.49) & 53.39 (-0.30)\\
& ${L_\infty }$ & - & - & - & 34.15 (-0.94) & 53.27 (-0.42)\\
& Cosine & - & - & - & 34.64 (-0.45) & 53.27 (-0.42)\\
& KL-Div. & - & - & - & 34.62 (-0.47) & 53.56 (-0.13) \\
\hline
\multirow{2}*{(B)} & - & Block & - & & 34.57 (-0.52) & 53.27 (-0.42)\\
& - &  Layer & - & - & 34.85 (-0.24) & 53.25 (-0.44) \\
\hline
\multirow{2}*{(C)} & - &  - & VGG19 & - & 34.38 (-0.71) & 53.24 (-0.45)\\
& - &  - & AlexNet & - & 33.98 (-1.11) & 52.99 (-0.70) \\
\hline
\multirow{4}*{(D)} & - & - & - & w/o (IrM-KD+IaM-KD) loss & 27.30 (-7.79) & 51.11 (-2.58)\\
& - & - & - & Image Space loss & 32.91 (-2.18) & 52.41 (-1.28)\\
& - & - & - & w/o IaM-KD loss & 33.64 (-1.45) & 52.81 (-0.88)\\
& - & - & - & w/o IrM-KD loss & 34.03 (-1.06) & 53.08 (-0.61)\\
\hline
\hline
\end{tabular}}
\end{table*}

\begin{table*}[!t]
\centering
\caption{Validation ablation results for diverse distillation variants on the EN-DE task.
The \textit{base} row denotes the IKD-MMT in Multi30K development sets, and ``-'' means to retain the setting of the \textit{base} row.
Dev.B and Dev.M indicate the BLEU and METEOR scores of the development set}
\label{table_VI}
\centering
\renewcommand{\arraystretch}{0.9} 
\setlength{\tabcolsep}{2.5pt}
{\begin{tabular}{c|cccc|cc}
\hline
\hline
& \textbf{Sim. Func.} & \textbf{Dist. Gran.} & \textbf{CNN Back.} & \textbf{Dist. Loss} & Dev.B & Dev.M \\
\hline
\textit{base} & $L_2$ & Model & ResNet50 & (IrM-KD+IaM-KD) loss & \textbf{42.48} & \textbf{59.20} \\
\hline
\multirow{4}*{(A)} & $L_1$ & - & - & - & 41.33(-1.15) & 58.55(-0.65)\\
& ${L_\infty }$ & - & - & - & 41.80(-0.68) & 58.92(-0.28) \\
& Cosine & - & - & - & 41.44(-1.04) & 58.64(-0.56)\\
& KL-Div. & - & - & - & 41.90 (-0.58) & 58.89(-0.31) \\
\hline
\multirow{2}*{(B)} & - & Block & - & & 41.83(-0.65) & 59.01(-0.19) \\
& - &  Layer & - & - & 41.67(-0.81) & 58.69(-0.51) \\
\hline
\multirow{2}*{(C)} & - &  - & VGG19 & - & 41.69(-0.79) & 58.80(-0.40) \\
& - &  - & AlexNet & - & 41.20(-1.28) & 58.45(-0.75) \\
\hline
\multirow{4}*{(D)} & - & - & - & w/o (IrM-KD+IaM-KD) loss & 36.02(-6.46) & 55.67(-3.53) \\
& - & - & - & Image Space loss & 40.45(-2.03) & 57.96(-1.24) \\
& - & - & - & w/o IaM-KD loss & 41.14(-1.34) & 58.39(-0.81) \\
& - & - & - & w/o IrM-KD loss & 41.46(-1.02) & 58.66(-0.54) \\
\hline
\hline
\end{tabular}}
\end{table*}

\subsection{Ablation Studies}
Table \ref{table_II} illustrates ablation experiments on the EN-DE task to explore the impact of different collocations of distillation modules.

\textbf{Similarity Function}
First, we explore the effect of using varied similarity functions to measure the divergence between hidden representations in our distillation module.
As shown in row (A), the ${L_2}$ norm is the best option.
Later, the performance order is KL Divergence \cite{kullback1951information} > ${L_1}$ norm > Cosine similarity > ${L_\infty }$.

\textbf{Distillation Granularity}
Second, in row (B), we analyze what distillation granularity would be the golden standard of our model for optimal translation performance.
Specifically, the ``Layer", ``Block" and ``Model" represents that we employ representations of each layer, each block, the last convolutional layer and the image in teacher-student models to compute the distillation loss.
Based on the evaluation results, we conclude that the ``Model'' is optimal, and the ``Block'' is consistent with the ``Layer'' in the Meteor score, but slightly inferior in the BLEU score.
The such phenomenon reflect that the initial and terminal representations in our knowledge distillation are sufficient to teach the student model to generate information-rich features. This case breaks the stereotype that KD must transmit all knowledge.

\textbf{CNN Backbone}
Third, in row (C), we devise three variants with diverse CNN backbones to investigate their impact on the translation.
The ResNet50 wins this round since the deep residual network can derive the strongest visual representation.
The VGG19 performs worse with the absence of residual connection and plenty of training samples for model convergence.
Undoubtedly, the lightweight AlexNet incurs the worst translation degradation.
It implies that the feature extraction capability of a small model may be difficult to undertake the heavy task of multi-supervised learning.

\textbf{Distillation Loss}
Finally, we discuss the translation performance of different distillation loss strategies in row (D).
Unsurprisingly, w/o (IrM-KD+IaM-KD) loss suffers the severest performance degradation.
Removing visual distillation leads to the absence of visual perception in multimodal features, which evolves into a perturbed feature obtained by passing the global text feature into the Fc$\&$Avg Unpool.
Afterwards, w/o IrM-KD loss outperforms w/o IaM-KD loss, indicating that the capability of the IaM-KD to establish the text-image relevance that is critical for multimodal feature synthesis is stronger than the IrM-KD.
We assume this event is related to that the IaM-KD covers the propagation path of the IrM-KD.
Compared with Image Space loss, the improvement of our method reveals that the intermediate hidden state of the teacher model plays a vital role in teaching the student model to comprehend the text-image correlations, as also verified in preceding KD work \cite{romero2014fitnets,yim2017gift}.
Overall, each distillation loss considerably improves translation.

\textbf{Ablation Studies on Development set}
Table \ref{table_VI} attaches all the validation ablation results to corroborate that each distillation hyperparameter also contributes its decent gains on the model convergence rather than just the model generalization.
Drawing from the tabular results, all hyperparameters can be tuned freely on the dev set. We further notice that the performances on the dev set align quite well with the testing set, in terms of tendency.

\section{Analysis}
In this section, we will investigate our IKD-MMT model from multiple perspectives.

\subsection{Does IKD-MMT really generate multimodal features?}
To explore the multimodal features generated by our distillation strategy, we test their informative richness from three aspects:

\begin{table}[!t]
\centering
\caption{Image retrieval tasks on the Multi30K dataset.}
\label{table_III}
\centering
\renewcommand{\arraystretch}{0.6} 
{\begin{tabular}{c|cccc}
\hline
\hline
& R@1 & R@5 & R@10 & R@15 \\
\hline
Train & 0 & 0.02 & 0.04 & 0.05 \\
Valid & 0.1 & 0.69 & 0.89 & 1.38 \\
Test2016 & 0.1 & 0.7 & 1.0 & 1.5 \\
Test2017 & 0.1 & 0.5 & 1.1 & 1.5 \\
MSCOCO & 0.22 & 0.65 & 2.39 & 2.82 \\
\hline
\hline
\end{tabular}}
\end{table}

\textbf{Image Retrieval}
The image retrieval task aims to analyzes the relationship between our generated multimodal feature and the visual feature.
Specifically, we generate the multimodal feature from each source sentence.
Further, we find the K closest visual features for each multimodal feature based on cosine similarity.
Then, we measure the R@K score, which calculates the recall rate of the visual feature of current sample in these top K nearest neighborhoods.
The results in Table \ref{table_III} display that no matter any K, or whichever data set, the R@K scores are extremely low.
These retrieval scores confirm that our model is not trying to generate the visual feature of the current image.

\begin{figure}[!t]
\centering
\setlength{\abovecaptionskip}{-3pt}
\setlength{\belowcaptionskip}{-10pt}
{\includegraphics[width=1.\linewidth]{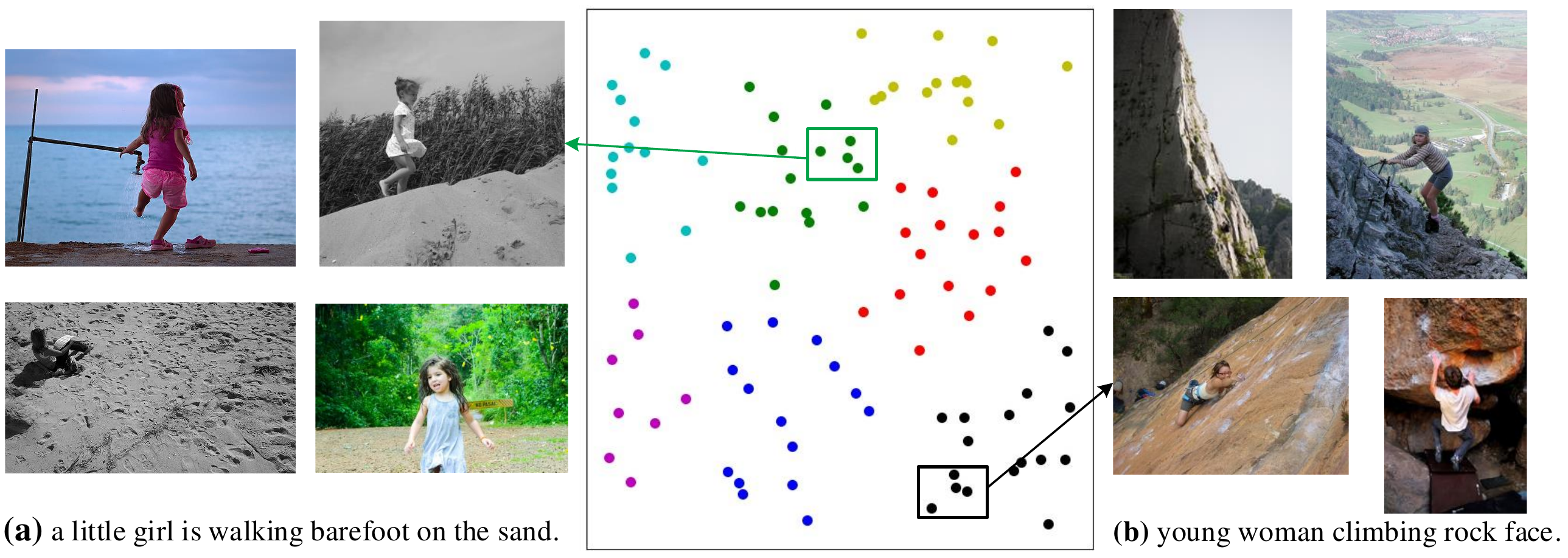} 
\label{fig3}}
\centering
\caption{The cluster analysis of the learned multimodal feature, where the two colored boxes represent some representative images in the two cluster cases. The arrow points to the original image that is belonged to the current multimodal feature (i.e. cluster center).}
\label{fig:Fig3}
\end{figure}

\textbf{Cluster Visualization}
In Figure \ref{fig:Fig3}, we visualize the related pictures which retrieved by the multimodal feature at the cluster map.
Here, points of different colors fall into different clusters, and the distance between points is specified by the cosine similarity between multimodal features and visual features.
In the cluster case (a), the other images exist the points-of-parity with the original image, namely objects, backgrounds, and actions (girl, sand, walking). Likewise, in the cluster case (b), the other images satisfy the identical thematic content (person, rock, climbing) as the original image.
Certainly, these related pictures also conform to the original text's description of the scene.
So the multimodal features are confirmed to have learned commonalities between images.

\begin{figure}[!t]
\centering
\setlength{\abovecaptionskip}{-1pt}
{\includegraphics[width=1.\linewidth]{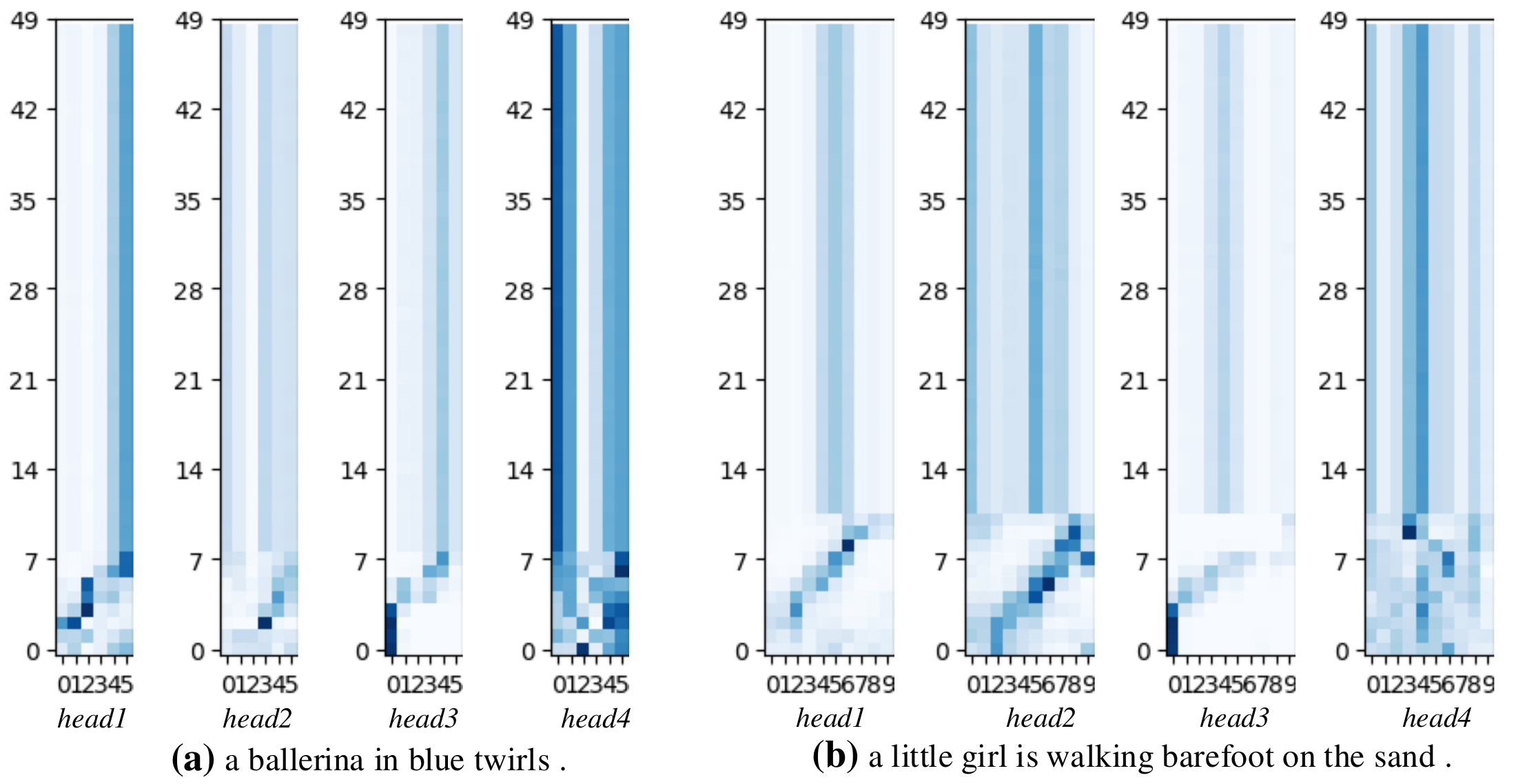} 
\label{fig4}}
\centering
\caption{Visualization of attention weights for fusion of multimodal features and text features. The weight values decreasing as the color becomes lighter.}
\label{fig:Fig4}
\end{figure}

\begin{figure*}[!t]
\centering
\setlength{\abovecaptionskip}{-0.5pt}
{\includegraphics[width=0.9\linewidth]{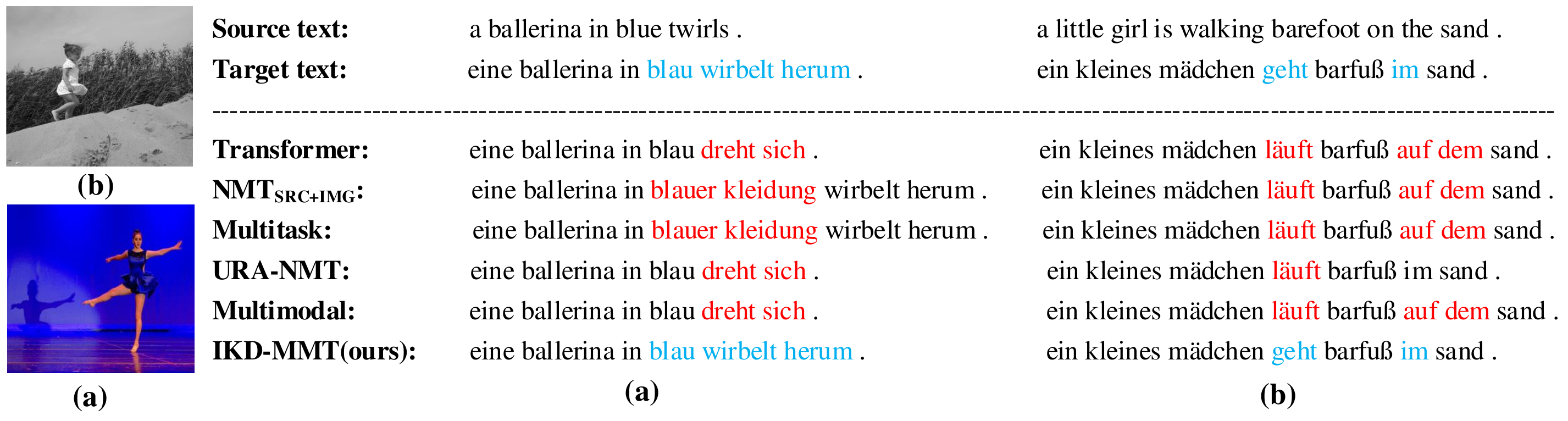} 
\label{fig5}}
\centering
\caption{Translation cases of different models. The red and blue highlight error and correct translations respectively.}
\label{fig:Fig5}
\end{figure*}

\textbf{Attention Weights}
In Figure \ref{fig:Fig4}, we envision the attention weights\footnote{They are computed from the 4 attention heads in the first multimodal transformer encoder layer.} for the fusion of multimodal features and text features.
These weights display which text words the different regions of the multimodal feature focus on.
Combining the two examples, several insights are excavated as follows:
1) Part of the multimodal feature with the size of sentence length can be regarded as "pseudo-words", and a word alignment is formed with the text feature.
2) The rest of the multimodal features pay the attention to words equally. We conjecture that these regions as non-object parts thus tend to contribute a consistent impact on text translation.
3) The attention weight of the former three attention heads are flat and presents linearization at the bottom part, while one of 4th attention head is fluctuating and presents dispersion at the bottom part.
This means that the first three attention heads capture the entire sentence semantics with the "global attention" form.
The 4th attention head, acts like the "local attention", and emphasizes understanding the keywords of the sentence.
These findings demonstrate that the multimodal features have embedded the textual semantics.

To summarize, the above experiments can fully prove that our IKD-MMT reliably generates an information-rich multimodal feature.

\subsection{Can multimodal features be directly used for translation?}
Our IKD-MMT model synthesizes a multimodal feature equipped with textual and image knowledge through a multimodal generator.
A natural question to ask is, can multimodal features be fed into the encoder alone, rather than being cascaded with textual features for translation?

To this end, we compare the model removing text features with the original benchmark in Table \ref{table_IV}.
We notice that w/o Text Feat. appears a cliff-like performance drop, which is explainable.
In the multimodal encoding layer, the dot product of the query and key vectors is used to mark the importance of each token corresponding to other tokens in the sentence, i.e. the attention score.
If we treat the multi-modal feature as the query, its fixed $P$ regions can be regarded as a set of pseudo tokens.
Considering this token set carries limited semantics and destroys the word alignment, it is difficult to obtain an available attention score alone.
In addition, most studies convey that text semantics is more important than visual perception in the MMT task \cite{gronroos-etal-2018-memad, lala-etal-2018-sheffield}.

\subsection{Can multimodal features recover the missing text?}
\begin{table}[!t]
\centering
\renewcommand{\arraystretch}{1.} 
\caption{Results of IKD-MMT without text features.}
\label{table_IV}
\centering
\setlength{\tabcolsep}{0.55pt}
{\begin{tabular}{l|cccccc}
\hline
\hline
\multicolumn{1}{l|}{\multirow{2}*{}} &  \multicolumn{2}{c}{\textbf{Test2016}} & \multicolumn{2}{c}{\textbf{Test2017}} & \multicolumn{2}{c}{\textbf{MSCOCO}}\\
\cline{2-7}
& B & M & B & M & B & M\\
\hline
IKD-MMT & 41.28 & 58.93 & 33.83 & 53.21 & 30.17 & 48.93 \\
\ \ w/o Text Feat. & 22.06 & 39.44 & 19.35 & 36.67 & 16.00 & 32.54\\
\hline
\hline
\end{tabular}}
\end{table}

\begin{table}[!t]
\centering
\renewcommand{\arraystretch}{1.} 
\caption{Results of two degraded text and original text\footnote{Here, we use the result of the reproduced models.}.}
\label{table_V}
\centering
\setlength{\tabcolsep}{0.3mm}
{\begin{tabular}{ccccc}
\hline
\hline
Model & $\mathcal{D}$ & $\mathcal{D{_C}}$ & $\mathcal{D{_E}}$\\
\hline
Transformer & 52.5 & 50.28 ($\downarrow$2.22) & 33.81 ($\downarrow$ 18.61) \\
Multimodal & 53.18 & 51.30 ($\downarrow$1.88) & 35.04 ($\downarrow$ 18.14) \\
IKD-MMT & 53.21 & 51.30 ($\downarrow$1.91) & 34.59 ($\downarrow$ 18.62) \\
\hline
\hline
\end{tabular}}
\end{table}

In Table \ref{table_V}, we adopt two degradation strategies \cite{ive-etal-2019-distilling, caglayan-etal-2019-probing} for the source sentence, and feed into Transformer, Multimodal and our IKD-MMT, to probe whether multimodal features can recover the missing text.
Test2017 Meteor scores are used for evaluation.

\textbf{Color Deprivation}
We mask the source tokens that refer to colors as a special token [U], which involves 3.19\% and 3.16\% of the words in the training set and test set, respectively.
As shown in the column $\mathcal{D{_C}}$, after color deprivation, the text-only Transformer fails to align the source and target tokens, then leads to the worst performance descent.
Our IKD-MMT and Image-must Multimodal hardly synthesize color information to compensate for the deterioration of color missing.

\textbf{Entity Masking}
We tag all visually depictable entities \cite{plummer2015flickr30k} with a special token [U], which affects 29.49\% and 31.12\% of the words in the training set and test set, respectively.
In the $\mathcal{D{_E}}$ column, we observe that the IKD-MMT and the text-only Transformer degrade equally in performance, which is because IKD-MMT unable to distill the visual perception of multimodal features from the entity-missed text.

Beyond these two masking experiments, we revisit such token recovery problems to pose a more common-sense insight:
As per the faithfulness-first principle \cite{koehn2009statistical} in translation, once the source sentence misses keyword information, what we need to do is translate this degraded faithfully. Re-translate back to the original target text from the degraded source text is false.

\subsection{Case Study}
Figure \ref{fig:Fig5} depicts the 1-best translation of the two test cases generated by various systems.
Other systems mistranslate and over-translate text in case (a) and distort the semantics due to mistakenly translating "geht" (walking) to "läuft" (running) in case(b).
Our IKD-MMT relies on rich multimodal semantics to keep the translation fidelity.

\section{Conclusion}
In this work, we propose the IKD-MMT framework to address the image-must issue for multimodal machine translation (MMT) via the knowledge distillation paradigm.
Under this image-free MMT system, there are three key contributions:
1) An information-rich multimodal feature is generated by the dual constraints of visual distillation and text translation to support the image-free testing stage;
2) The knowledge distillation module is flexible, and pioneers to employ of the pre-trained model to guide translation;
3) Both quantitative and qualitative results validate the feasibility of the proposed approach IKD-MMT, where it can be deemed the first framework that rivals or even surpass most (if not all) image-must frameworks.

\section*{Acknowledgements}
Ru Peng and Junbo Zhao were supported by the Fundamental Research Funds for the Central Universities (No. 226-2022-00028).
Junbo Zhao also wants to thank the Zhejiang University startup package.
The authors would like to thank the Institute of Computer Innovation of Zhejiang University for the high-performance computing platform.

\section*{Limitations}
From a representation learning perspective, this work is dedicated to introduce the visual perception pipeline and the comprehension of text-image correlations from texts, and may be limited to more complex visual descriptive text (if there exists numerous visual descriptive entities).
Further, since the Multi30K is the only and most commonly used MMT benchmark, most of the experiments are centered around it.
We addtionally made a ``bold'' attempt to move IKD-MMT onto much larger scaled NMT datasets for testing only --- thanks to the image-free nature of our approach --- and unfortuantely the inference results did not look decent enough.
While the IKD-MMT's image-free inference pass can be fully facilitated in this scenario,
we attribute the inferior results to the much simpler data distribution involved in the Multi30K. Indeed, we hope a richer or real-world MMT dataset could fully bridge this image-free performance gap between MMT and NMT.
That, however, may have gone beyond the scope of this paper.

\bibliography{emnlp2022}
\bibliographystyle{emnlp2022.bib}

\appendix
\section{Appendix}
\label{sec:appendix}
\begin{table*}[hb]
\centering
\setlength{\tabcolsep}{2.5pt}
\caption{Architecture of each module in multimodal feature generation.
The multimodal student model has inverted data flow with visual teacher model.
These architectures can be easily replaced with any CNN variant (e.g. VGG19 \cite{simonyan2014very}, AlexNet \cite{krizhevsky2012imagenet}) with reference to ResNet50 \cite{he2016deep}.
}
\label{table_VII}
\centering
{\begin{tabular}{c|c|c||c|c|c}
\hline
\hline
\multicolumn{3}{c}{\textbf{Visual Teacher Model}} & \multicolumn{3}{c}{\textbf{Multimodal Student Model}}\\
\hline
layer name & output size & 49-layer & layer name & output size & 48-layer\\
\hline
T-conv1 & 112x112 & 7x7, 64, stride 2 & S-conv1 & 224x224 & 8x8, 3, stride 2 \\
\hline
\multirow{4}*{T-conv2\_x} & \multirow{4}*{56x56} & 3x3 max pool, stride 2 & \multirow{4}*{S-conv2\_x} & \multirow{4}*{112x112} & 2x2 max unpool, stride 2\\
\cline{3-3} \cline{6-6}
& & $\begin{bmatrix}
    1\text{x}1, 64\\
    3\text{x}3, 64\\
    1\text{x}1, 256\\
\end{bmatrix}\text{x}3$ &
& & $\begin{bmatrix}
    1\text{x}1, 256\\
    3\text{x}3, 64\\
    1\text{x}1, 64\\
\end{bmatrix}\text{x}3$
\\
\hline
T-conv3\_x & 28x28 & $\begin{bmatrix}
    1\text{x}1, 128\\
    3\text{x}3, 128\\
    1\text{x}1, 512\\
\end{bmatrix}\text{x}4$ &
S-conv3\_x & 56x56 & $\begin{bmatrix}
    1\text{x}1, 512\\
    3\text{x}3, 128\\
    1\text{x}1, 128\\
\end{bmatrix}\text{x}4$ \\
\hline
T-conv4\_x & 14x14 & $\begin{bmatrix}
    1\text{x}1, 256\\
    3\text{x}3, 256\\
    1\text{x}1, 1024\\
\end{bmatrix}\text{x}6$ &
S-conv4\_x & 28x28 & $\begin{bmatrix}
    1\text{x}1, 512\\
    3\text{x}3, 256\\
    1\text{x}1, 256\\
\end{bmatrix}\text{x}6$ \\
\hline
T-conv5\_x & 7x7 & $\begin{bmatrix}
    1\text{x}1, 512\\
    3\text{x}3, 512\\
    1\text{x}1, 2048\\
\end{bmatrix}\text{x}3$ &
\multirow{2}*{S-conv5\_x} & \multirow{2}*{14x14} & \multirow{2}*{$\begin{bmatrix}
    1\text{x}1, 1024\\
    3\text{x}3, 512\\
    1\text{x}1, 512\\
\end{bmatrix}\text{x}3$} \\
\cline{1-3}
N/A & 1x1 & average pool & & & \\
\hline
\hline
\multicolumn{3}{c||}{\textbf{Multimodal Feature Generator}} & N/A & 7x7 & 2048-d fc, average unpool \\
\hline
\hline
\end{tabular}}
\end{table*}

\end{document}